# Hair and Scalp Disease Detection using Machine Learning and Image Processing

Mrinmoy Roy and Anica Tasnim Protity

## ABSTRACT

Almost 80 million Americans suffer from hair loss due to aging, stress, medication, or genetic makeup. Hair and scalp-related diseases often go unnoticed in the beginning. Sometimes, a patient cannot differentiate between hair loss and regular hair fall. Diagnosing hair-related diseases is time-consuming as it requires professional dermatologists to perform visual and medical tests. Because of that, the overall diagnosis gets delayed, which worsens the severity of the illness. Due to the image-processing ability, neural network-based applications are used in various sectors, especially healthcare and health informatics, to predict deadly diseases like cancers and tumors. These applications assist clinicians and patients and provide an initial insight into early-stage symptoms. In this study, we used a deep learning approach that successfully predicts three main types of hair loss and scalp-related diseases: alopecia, psoriasis, and folliculitis. However, limited study in this area, unavailability of a proper dataset, and degree of variety among the images scattered over the internet made the task challenging. 150 images were obtained from various sources and then preprocessed by denoising, image equalization, enhancement, and data balancing, thereby minimizing the error rate. After feeding the processed data into the 2D convolutional neural network (CNN) model, we obtained overall training accuracy of 96.2%, with a validation accuracy of 91.1%. The precision and recall score of alopecia, psoriasis, and folliculitis are 0.895, 0.846, and 1.0, respectively. We also created a dataset of the scalp images for future prospective researchers.

M. Roy
Department of Computer Science, Northern Illinois University, USA.
(e-mail: mrinmoy.cs10@gmail.com)
A. T. Protity
Department of Biological Sciences, Northern Illinois University, USA.
(e-mail: protity.microbiology@gmail.com)

*Corresponding Author

## I. Introduction

Hair, made of keratin protein, pertains to beauty and masculinity. Approximately 5 million hair follicles are present throughout our body [1]. Scalp Hair maintains body temperature and protects the brain from external heat. A typical hair growth cycle runs for 2-7 years, according to [2] and [3]. A healthy human has 100,000 hairs on the scalp, and 50-100 hair loss per day is considered normal. Hair loss is not a present-day issue. The hair-loss treatment was found in ancient Ayurveda scriptures 6000 years ago [2]. However, Hair and scalp-related issues are gaining more recognition nowadays compared to earlier years due to certain factors, such as environmental pollution, hormonal imbalance, autoimmune disease, gut microbiota alteration, elevated physical and mental stress levels in human lifestyle, seasonal change, unhealthy diet, micronutrient deficiency, genetic predisposition, and side-effects of drugs [2], [3]. According to [4], 80 million Americans have hair loss-related issues to some extent. Although most hair loss diseases are localized, some can spread to other locations. Some diseases require prescribed drugs and hair transplantation. Some diseases are caused by bacterial or fungal infections and require antibiotic treatment. Often, there are genetic and sexual predispositions in hair-scalp diseases.

Alopecia, folliculitis, and psoriasis are some common causes of hair loss. There is a difference between regular hair fall and alopecia; the latter develops coin-sized bald patches all over the scalp area. Alopecia or patchy hair loss can be of different types. Androgenetic alopecia or male-pattern baldness (MPB) is the most common form of alopecia where the hairline starts to recede, following a pattern where the frontal and temple area are most affected. 70% of men and 40% of women get this type of hair loss and thinning issue [3]. According to [5], MPB is an X-linked polygenic disease, and males are more genetically prone to develop baldness at a mature age. Topical minoxidil solution thickens the hair by 50% [3]. On the other hand, Alopecia areata (AA) is an autoimmune disease affecting individuals irrespective of age and sex. Primarily affecting the scalp area, AA can also spread in the beard, eyelashes, and eyebrows. In this case, the body's immune cells cannot recognize hair follicles as 'self.' Instead, they consider these follicles as 'foreign,' which ultimately causes the hair follicles to be targeted and destroyed by the immune cells. It is an example of a hereditary disease. The study from [6] reported that, in the US alone, 700,000 individuals suffer from AA. This disease, if diagnosed early, might resolve spontaneously. In severe cases, topical corticosteroid or immune therapy is used [3].

Sometimes, the hair follicles might get inflamed because of the action of bacterial accumulation. This follicle inflammation is called folliculitis decalvans. The bacterium Staphylococcus aureus damages the follicle and prevents hair growth. Staphylococcus aureus uses hair tufts to enter underneath the follicle, causing chronic inflammation, redness, swelling, scarring, itching, and hair loss. Antibiotic treatment combined with surgical removal of hair tufts and corticosteroids for reducing inflammation are the prescribed treatment for Folliculitis decalvans [3]. Psoriasis is another form of common scalp skin disease. According to [7], 54% of 5600 psoriasis patients had scalp psoriasis. Severe psoriasis may cause significant itching, scaling, and redness in the scalp. The application of topical shampoo and corticosteroids are the treatment options by [8].

Some scalp infections may be treatable if diagnosed early. Some but not all diseases may go on their own. Only an expert physician can detect the illness by visual observation. In some cases, early disease detection is beneficial for dermatologists to initiate the treatment. An early scalp inspection includes a dermatoscopic examination of the scalp for inflammation, itching, localized lesion, dandruff, follicular flakes, louse eggs (nits), and a scalp biopsy. Besides visual observation, the patient can undergo blood and hormone tests to detect the exact disease. Unfortunately, most hair and scalp diseases are diagnosed in advanced stages, which complicate the treatment options. All these factors lengthen the diagnosis and treatment process. Therefore, researchers are putting more effort into developing different mechanisms for the early detection of hair and scalp diseases.

In the 21st century, with all the advancements in computational technology, extensive application of machine learning has made our daily lives simple, comfortable, and secure. The increasing popularity of machine learning and its nature to extract patterns from data are directing researchers to incorporate several machine learning algorithms into health informatics. Especially during the Covid-19 pandemic era, different applications like restraining people from covid-19 spread [9], SARS-CoV-2 screening and treatment [10], lock-down control in case of high dimensional input [11] came into play, which made machine learning and healthcare systems inseparable. Overall, adapting, integrating, and developing deep learning-based applications on patients' information, medical reports, and audio-video feedback make the diagnosis process faster. Nowadays, patients can get at least the initial idea of disease detection by themselves using easily accessible smart devices. All these applications clear their confusion and help them make health-related decisions independently.

The high computational capability of neural networks is, therefore, a breakthrough in healthcare and medical diagnostic organizations. Convolutional neural networks (CNN) have brought revolutionary success in detecting deadly diseases. To date, neural networks are assisting healthcare professionals in the early detection of different types of tumors and cancers, such as skin cancer (melanoma) [12], stomach cancer (adenocarcinoma) [13], and brain tumors (glioblastoma) [14]. Neural networks are applicable in detecting life-threatening dengue fever [15] and Covid-19 [16] as well. In one study, CNN was used to extract complex temporal dynamic features from heart rate variability (HRV) signals, developing an algorithm that facilitated the early detection of diabetics [17]. Using the image processing ability of the neural networks, we can extract features from hair, skin and scalp images to classify and categorize numerous hair and scalp-related diseases. In this work, due to the importance of early-stage hair disease detection, we applied convolutional neural networks to 3 types of hair diseases and developed a model to detect them successfully.

## II. CHALLENGES AND CONTRIBUTIONS

A classic application of computer vision is to detect disease using digital images. Researchers can exploit a pool of digital images obtained from one or more datasets, preprocess the images, feed the images into the neural network, and develop a model to detect the disease. Unfortunately, minimal research has been performed on the machine-learning approach for scalp disease detection. There are several unique challenges behind this. First and foremost, hair diseases are not localized and can spread to different regions of the scalp, beard, eyebrows, eyelashes, and pubic area. Second, every image needs different types of preprocessing before feeding to neural networks. Different scalp skin tones, hair colors, and types around the detection zones make the imaging process more complicated. Third, no proper dataset for scalp diseases is available over the internet, and images taken from the internet differ in size and resolution. Moreover, one must be conscious of minimalizing and correcting the error in disease detection; otherwise, the high false-positive and false-negative rates result in misdiagnosis of the disease and worsening hair loss.

To overcome the challenges, we developed a model which can successfully classify the alopecia, folliculitis, and psoriasis diseases with a minimal false-positive and false-negative rate. Though it is challenging to collect images for the diseases from the internet, and the images are varied in color, shape, and resolution, we applied various preprocessing, such as denoising, resizing, enhancement and created a dataset that might help in further scalp diseases research.

## III. RELATED WORKS

Disease detection using machine learning approaches is gaining popularity in health informatics. Many skin and scalp-related diseases can be detected using images of infected regions within a few seconds. In one study by [18], a framework is developed to differentiate alopecia areata from healthy hair. They obtained 200 healthy hair images from the figaro1K dataset and 68 alopecia areata hair images from DermNet. After a series of enhancement and segmentation, three key features were extracted from the images: texture, shape, and color. The researchers divided the dataset into 70%-30% train-test-split and applied a support vector machine (SNM) and k-nearest neighbor (KNN) for the classification task. Overall, they achieved 91.4% and 88.9% accuracy using SVM and KNN, respectively, with a 10-fold cross-validation approach. However, using other machine learning algorithms might increase in the accuracy rate, which should have been discussed. Besides, the application of Histogram Equalization (HE) for image enhancement complicated the process of getting accurate texture features from distorted images, as HE itself adds noise to the output image, distorting the signals. Moreover, this study only shed light on alopecia areata disease, ignoring the inter-class differences between other similar type diseases, which increased the likelihood of inaccurate prediction of other diseases as alopecia areata, thereby making this framework less reliable.

Another study [19] proposed a model for early alopecia detection. They used 100 samples for this research, with 80% as training data and the other 20% as testing data. They looked for four attributes, length of the hair, nail brittleness, amount of damage made to the hair, and hair follicle. Two-layer feed-forward network with a back propagation technique was used for detection purposes. The proposed model system consisting of 4 input neurons, 10 hidden neurons, and a linear output neuron, achieved 91% training accuracy with 86.7% validation accuracy. It showed the best performance at epoch 4 with a 0.059687 gradient. However, the study has some pitfalls, too, as they did not mention their data source or differentiate data classes with their respective sample sizes. Also, no image pre-processing was performed on the collected images. Although there is a possibility of overfitting without a proper data balancing technique, this report did not discuss the data balancing between the two classes. Furthermore, they did not calculate the model's false-positive and false-negative rates, which is crucial for a model specially developed for the healthcare system.

Related work [20] was performed on skin disease detection, where machine learning was used to analyze the digital image of the affected skin area for identifying eczema, melanoma, and psoriasis. Their dataset consists of 80 images from different websites specific to skin diseases. By using a convolutional neural network for feature extraction and applying multiclass SVM on those features, they achieved 100% accuracy in disease classification. However, they did not explore other essential model performance matrices and overfitting issues. In another skin disease detection-based article [21], the authors proposed a scheme to classify skin lesions into five categories: healthy, acne, eczema, benign, and malignant melanoma, using a pre-trained CNN model, AlexNET for feature extraction and error correcting output codes support vector machine for classification. The dataset consists of 9144 images from different sources and achieved 84.21% accuracy using a 10-fold cross-validation technique.

Overall, we observed very few works on hair diseases. The recent related works lack at least one of the following categories – discussion over false positive and false negative rates, ignoring inter-class differences, model reliability, and overfitting problem. In this work, we have attempted to fill these gaps by leveraging a convolutional neural network algorithm on hair disease images while maintaining high accuracy with good precision and recall scores.

## IV. Data Description & Device

### A. Data Collection

The most challenging part of using visual images for disease prediction and disease classification is data collection. Often, one can get fewer appropriate images for a specific illness found. Moreover, the pictures are scattered over the internet. In this study, the authors extracted the images from different websites, such as DermQuest, DermNet, MedicineNet, DermnetNZ, and various medical professionals.

TABLE I: Images per Disease

| Disease | Quantity |
|---|---|
| Alopecia | 65 |
| Psoriasis | 45 |
| Folliculitis | 40 |

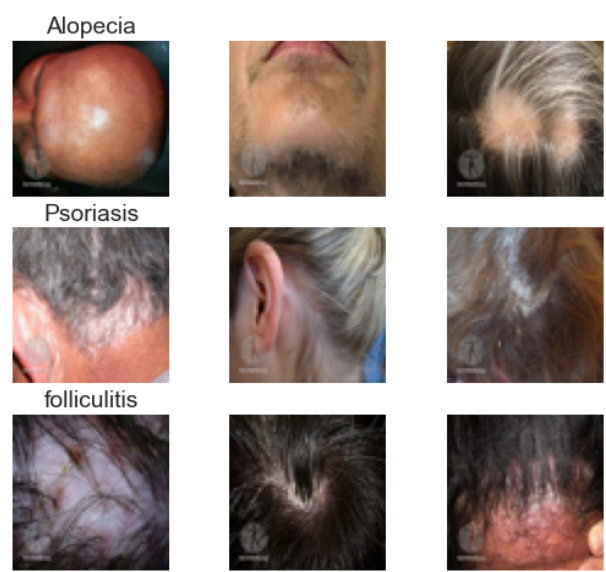

Fig. 1. Image subset of each disease category.

The image quantity is different for each category. We found more alopecia-related images than other diseases because alopecia is more frequent and severe among the human population. The number of samples in each type of disease is listed in Table I. Randomly selected images in each category are graphically represented in Fig 1. Our dataset is made publicly available on [22].

### B. Device

The research was conducted on Dell Latitude 5520 laptop device having 11th generation Intel Core i5 (8 MB cache, 4 cores, 8 threads, up to 4.40 GHz Turbo) and running on Windows 10 Pro operating system. The device has 16 GB, 1 x 16 GB, DDR4, 3200 MHz random access memory (RAM), and 256 GB, M.2 PCIe NVMe, SSD, Class 35 (NVRAM). For the classification of images, we utilized the integrated Intel Iris XE graphics capable with a thunderbolt for I5-1145G7 vPro processor. For the data collection, we used iPhone-13 Pro Max having Hexa-core (2x3.23 GHz Avalanche + 4x1.82 GHz Blizzard) CPU and Apple GPU (5-core graphics). We used a mobile device with 128GB 6GB RAM, and a 12 MP triple main camera for the image collection.

## V. Proposed Model

In this section, we introduce the system workflow of our model and explain the functions of each module in details. As shown in Fig. 2, first, the captured image is sent to preprocessing steps which are divided into three parts: image equalization, image enhancement, and data balancing.

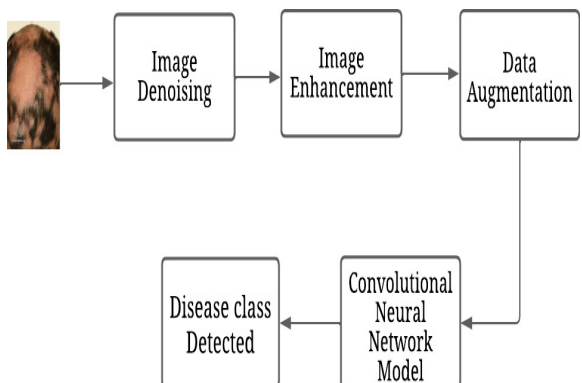

Fig. 2. System workflow of hair disease detection model.

Among these three, the first two parts are mainly for increasing image quality, and the last part is for model versatility. After the preprocessing steps, the image is passed to the Neural Network model for the classification task. We used a convolutional neural network that classifies an image successfully into three different classes: alopecia, folliculitis, and psoriasis.

### A. Denoising

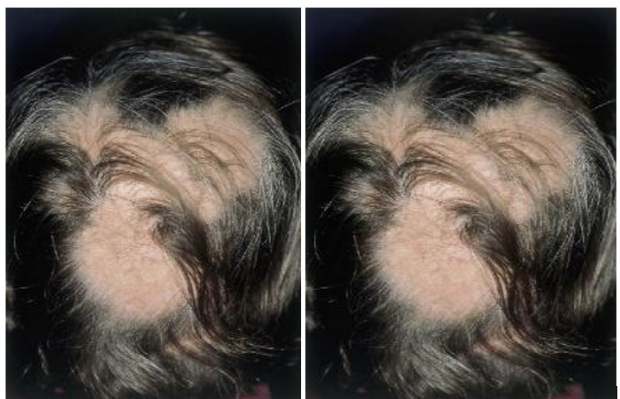

Fig. 3. Left original image & right non-local means denoised image.

Noise is the degradation of image signals caused by external sources [23]. Noise introduces random variations of brightness or color information in the captured images. Most of the time, images on the internet have some noise associated with them. As we have collected most of the data samples from different dermatology websites, the noise in our dataset is not homogeneously distributed, which made it more complex. Therefore, we applied additional filters for denoising the collected images. We started with the gaussian filter for a better image classification process. However, after using the gaussian filter, the images become completely blurred, which leads to the loss of important information and damage to the edges. We then applied the median filter, which worked better than the gaussian filter with kernel_size = 3. Though we achieved better accuracy using the bilateral filter, we got the best results while applying the non-local means filter with patch_size = 3 and patch_distance = 5. This non-local means filter preserved all the edges and reduced the noise better than the other filters for our application which is shown in Fig. 3.

### B. Image Equalization

Often the captured image doesn't reflect the natural view and needs contrast enhancement to meet the level of realistic view [24]. Especially images with high color depth and after denoising effects need normalization for a better realistic view [25]. First, we applied histogram equalization (HE). However, the HE increases the contrast of the background when used in images with low color depth, and information is lost as the histogram is not confined to the local region. To overcome the problem, we applied CLAHE (Contrast Limited Adaptive Histogram Equalization) by dividing an image into equal size non-overlapping areas and computing a histogram for each region. After clipping the histogram, we distributed the clipped value over the histogram equalization, which gives us control of the over-amplification of the contrast and generates the resultant image shown in Fig. 4.

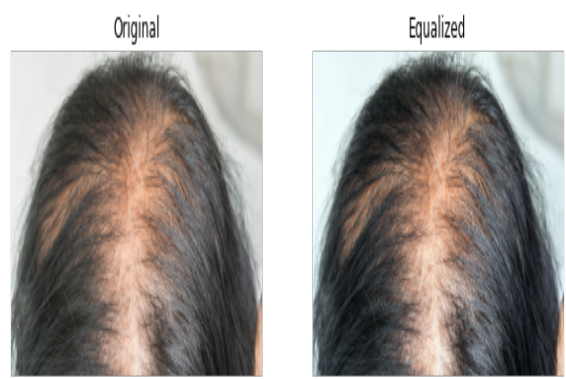

Fig. 4. Image equalization using CLAHE.

### C. Data Balancing

The overall performance of a machine learning model depends on the balanced dataset because, without it, minority class detection becomes difficult. Balancing a dataset reduces the risk of skewing towards the majority. Imbalanced data might achieve high accuracy, but the results are biased toward the majority class. As alopecia is a common disease, we have more alopecia images than other diseases, which creates an imbalanced dataset for our model. For balancing the dataset, we used data augmentation techniques (re-scaling, random rotating, cropping, vertical and horizontal flipping) and oversampled the infrequent class.

### D. Neural Network Model

Neural network is the most applied model for visual data analysis. Neural network needs limited human assistance and can identify complex non-linear relationship between input and output. From global or local scale modeling [26] to diagnosis by medical image classification, neural network is using extensively. Moreover, Facial recognition, image labeling, accurate video subtitles, assisting call centers, automated virtual agents all these things are using neural network. There are 3 types of neural network available: Artificial Neural Networks (ANN), Convolution Neural Networks (CNN) and Recurrent Neural Networks (RNN). Each neural network has mainly 3 components: an input layer, a processing layer, and an output layer.

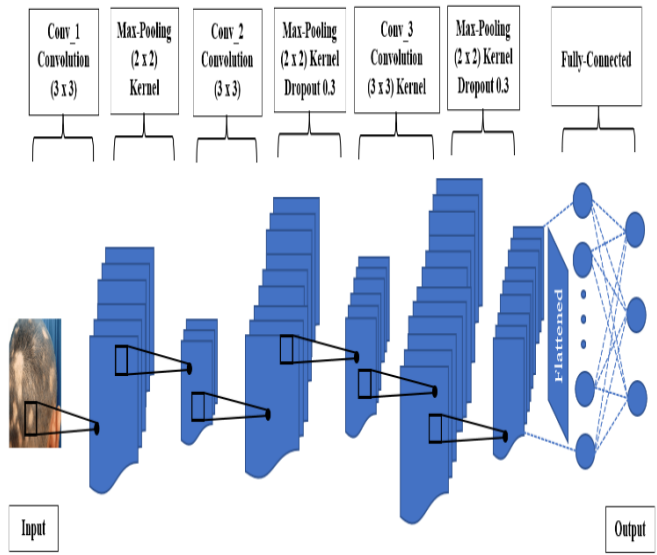

Fig. 5. Neural Network Model.

In this study, CNN is utilized for classification because it takes image's raw pixel data, trains a model, and extracts the features automatically for better detection. We used autokeras to find the best model for this problem. After trying 25 different combinations, we selected 3 hidden layers with 1 input and 1 output layer as our final model which is shown in Fig. 5. For training the model, we used batch_size = 16 with 50 epochs for each batch. The preprocessed data is divided into 70-30 train-test-split for training and validation purpose. Our model consists of 256 inputs, 3 x 3 square kernel, 3 output units and a softmax output. We used ReLU as our activation function to prevent the exponential growth of required computation and to explore the non-linear relationship between input and output variables. After each convolutional layer, input goes through the pooling layer having 2 x 2 kernel size to reduce the dimensions of the features map. Pooling layer summarizes the presented features in a region and helps to prevent the over-fitting problem by down sampling. We also used dropout layer after each pooling layer to prevent neurons in a layer from synchronously optimizing their weights and converging to the same goal. Our model's dropout rate is 0.3, which means 30% of the neurons of this layer will be randomly dropped in each epoch.

All the resultant 2-D arrays from pooled features map passes through the flatten layer and converted to single dimensional long continuous linear vector in the transition towards the fully connected layer as in Fig. 5. In the fully connected layer, every single output pixel from the convolutional layers is connected to 3 output classes. Though dense layer is computationally expensive, we used 2 dense layers for our classification task. Finally, we used softmax activation function to transform the 3 units of fully connected layer to a probability distribution which is represented by a vector of 3 elements, and the highest probability element selected as the final class. We leveraged adam optimizer for learning purpose and reducing the overall loss by changing the weights and learning rates. We used adam because it can handle sparse gradients on noisy problems and combines the best properties of AgaGrad and RMSProp algorithms.

## VI. RESULTS

We trained our CNN model using the optimal hyperparameters selected from the grid search. These hyperparameters are listed in Table II. We divided the dataset into 70%-30% train-test-split where 105 randomly selected images are used for training and 45 random images for testing. After applying the preprocessing steps, we used the training dataset to train the CNN model and evaluated the test dataset using the model.

TABLE II: HYPERPARAMETERS OF CNN MODEL

| Hyperparameters | Values |
|---|---|
| Batch Size | 16 |
| Epoch | 50 |
| Kernel Size | 3 x 3 |
| Optimizer | Adam |
| Dropout Rate | 0.3 |
| Pooling Size | 2 x 2 |
| Activation Function | ReLU |

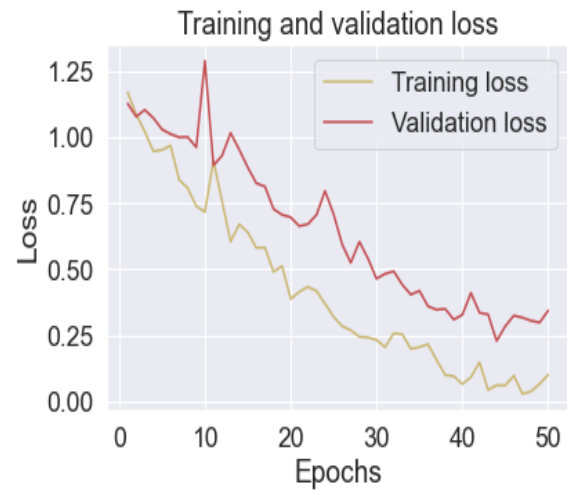

Fig. 6. Training and Validation loss for CNN.

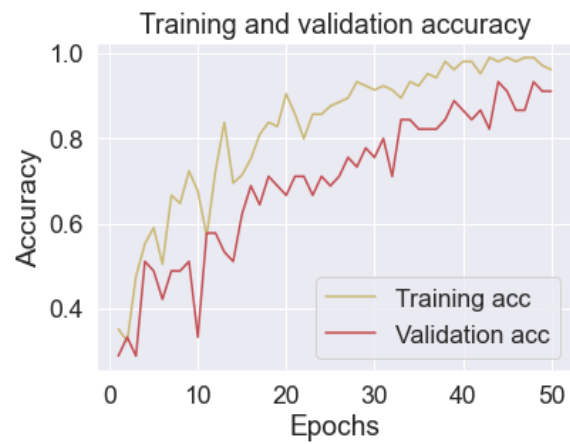

Fig. 7. Training and Validation Accuracy for CNN.

Our system achieved 96.2% training accuracy and 91.1% validation accuracy on the unseen data. Validation and training losses for every epoch are shown in Fig 6. The training losses decreased from 1.1685 to 0.1017, and the validation losses decrease from 1.1260 to 0.3438 while going from epoch 1 to epoch 50. At the same time, Training accuracy and validation accuracy increased to 96.2% and 91.1%, respectively, from epoch 1 to epoch 50, shown in Fig. 7.

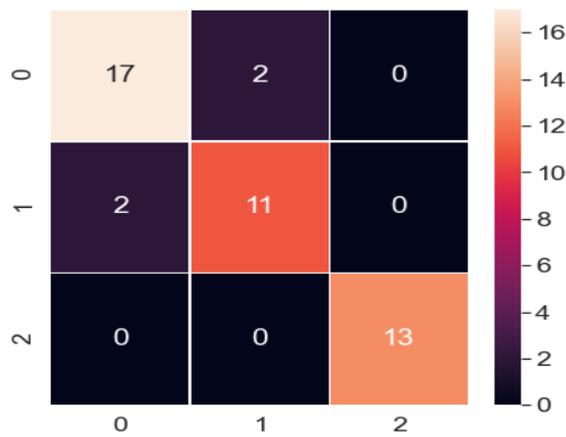

Fig. 8. Confusion Matrix of Our Model.

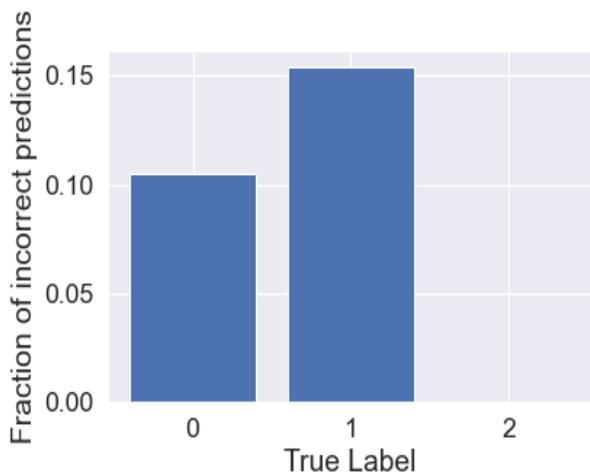

Fig. 9. Fractional Incorrect Prediction of Our Model.

The confusion matrix in Fig. 8 shows the correct and wrong classification for each category with inter-class classification. Among 45 test images, alopecia (label 0) has 19 images, psoriasis (label 1) has 13 images, and folliculitis (label 2) has 13 images. A total of 17 alopecia images were classified as alopecia and the other 2 were incorrectly classified as psoriasis. Again, 11 psoriasis images are classified as psoriasis, but 2 psoriasis images were incorrectly classified as alopecia. All 13 folliculitis images are classified correctly. The fractional incorrect prediction for each class is shown in Fig 9. Our model achieved the precision and recall score of 0.895 for the alopecia disease class, 0.846 for the psoriasis disease class, and 1.0 for the folliculitis disease class. As the precision and recall scores are same in each class, F1 scores are also similar to their respective precision and recall values.

## VII. Conclusion

Although early-stage detection of hair and scalp-related diseases is the key to the treatment process, hair loss and scalp diseases can often go undetected due to a lack of awareness and a lengthy diagnosis test. An AI-based application might pave the way to facilitate early disease detection. In this study, we developed a machine learning model to accurately predict three hair and scalp-related diseases: alopecia, folliculitis, and psoriasis by feeding 150 preprocessed image data into a 2-D convolutional neural network model. After using 70% of the data to train the model, we analyzed remaining 30% of images for testing our model. After subsequent training, the model gave an overall 96.2% training accuracy on the training data and 91.1% validation accuracy for the test data, with a high precision and recall scores for each disease type. We have also provided our dataset with this study. Our proposed system would assist dermatologists and patients with a better understanding of disease classification and initiating early treatment options for the three most frequently occurred hair and scalp diseases.

### Conflict of Interest

The authors declare that they do not have any conflicts of interest.